\documentclass[11pt]{article}

\usepackage{fullpage}
\usepackage{amsthm}
\usepackage{authblk}
\title{Polynomial-time Sparse Measure Recovery:\\ From Mean Field Theory to Algorithm Design}
\usepackage{times}
\usepackage{asymptote}
\usepackage{adjustbox}
\usepackage{multicol}
\usepackage{xfrac}
\usepackage{algorithm}
\usepackage{algpseudocode}
\usepackage[english]{babel}
\usepackage[margin=1in]{geometry}
\usepackage{amsmath, amssymb, bbm,  color}
\usepackage{pgfplots}
\pgfplotsset{compat=1.15}
\usepackage{mathrsfs}

\usetikzlibrary{arrows}
\usepackage{natbib}
\setcitestyle{square}
\usepackage{hyperref}
\usepackage{mathtools}
\DeclarePairedDelimiter\floor{\lfloor}{\rfloor}
\newcommand{\bigo}{\text{O}}
\newcommand{\R}{{{\mathbb {R}}}}
\newcommand{\C}{{{\mathbb C}}}

\renewcommand{\S}{{\mathcal S}}
\newcommand{\sign}{\text{sign}}
\newtheorem{Theorem}{Theorem}

\newtheorem {lemma} [Theorem]    {Lemma}

\newtheorem {Proposition}[Theorem]    {Proposition}
\newtheorem {theorem}[Theorem]    {Theorem}

\let\oldmarginpar\marginpar
\renewcommand\marginpar[1]{\-\oldmarginpar[\raggedleft\scriptsize #1]%
{\raggedright\scriptsize #1}}

\newcommand{\E}{\textbf{E}}

\newtheorem{assumption}{Assumption}



\begin{document}

% \title{\vspace{-1cm} polynomial-time Sparse Measure Recovery: Algorithm design based on Mean-field Wasserstein Gradient Flows}
\author[1]{Hadi Daneshmand\thanks{hdanesh@mit.edu}}

\author[2]{Francis Bach}
\affil[1]{Laboratory for Information and Decision Systems, MIT}
\affil[2]{INRIA-ENS-PSL Paris}
%  \author{\vspace{-0.1cm}Hadi Daneshmand}
%  \address{\footnotesize{MIT LIDS}}
% \author{\vspace{-0.2cm}Francis Bach}
% \address{\footnotesize{INRIA, Ecole Normale Sup\'erieure, Paris \\
% PSL University}}
% \email{hdanesh@mit.edu, francis.bach@inria.fr}

\maketitle

\begin{abstract}
   Mean field theory has provided theoretical insights into various algorithms by letting the problem size tend to infinity. We argue that the applications of mean-field theory go beyond theoretical insights as it can inspire the design of practical algorithms. Leveraging mean-field analyses in physics, we propose a novel algorithm for sparse measure recovery. For sparse measures over $\mathbb{R}$, we propose a polynomial-time recovery method from Fourier moments that improves upon convex relaxation methods in a specific parameter regime; then, we demonstrate the application of our results for the optimization of particular two-dimensional, single-layer neural networks in realizable settings. 
\end{abstract}
\vspace{1.cm}

\section{Introduction}

\subsection{Mean field theory}
Mean field theory has shown great promise in the analysis of algorithms widely used in machine learning, and computer science, including the optimization of shallow~\citep{bach2021gradient,chizat2021convergence} and deep neural networks~\citep{matthews2018gaussian,pennington2018emergence,pennington2017resurrecting,xiao2018dynamical}, and average case computational complexities for NP-hard problems~\citep{ding2015proof}. The mean field regime is a theoretical setting to study problems where the number of parameters (or input size) goes to infinity. Such \textit{overparameterization} reveals coarse structures by integrating fine details.

There is a gap between the mean-field regime and the regime of finite problem size~\citep{lewkowycz2020large,li2022neural,matthews2018gaussian}. Since mean-field predictions do not necessarily hold in standard settings~\citep{bach2021learning}, a recent line of research investigates how accurate the mean-field predictions are~\citep{li2022neural}. We use a different approach to fill the gap between the theoretical mean-field regime and the standard problem settings. We design an implementable (polynomial-time) algorithm using a particular mean-field analysis. This algorithm is proposed for sparse measure recovery, which is a general framework that encompasses important problems across various domains.

\subsection{Sparse measure recovery}
Consider a measure $\mu$ with a finite support:
\begin{align*} 
\mu = \frac{1}{n} \sum_{i=1}^n \delta_{w_i},
\end{align*}where $w_i \in  \Omega \subset \R^d$, and $\delta_{w_i}$ is the Dirac measure at $w_i$. Notably, the vector $w_i$ may represent an image or parameters of a neuron in neural networks, which are assumed to be distinct~\citep{candes2014towards,tang2014near}.  

% \begin{assumptioncost}{1}{\ell}
\begin{assumption}
\label{assume:distinct}
There exists a positive constant $\ell$ such that 
$
     \| w_i - w_j \| \geq \ell
$
holds for $i\neq j \in [n]$.
\end{assumption}
  Given a map $\Phi:\Omega \to \C^m$, the generalized moments of $\mu$ are defined as \citep{duval2015exact}
\begin{align} 
\label{eq:moments} \tag{generalized moments}
     \Phi \mu=  \int \Phi(w) d \mu(w) \in \C^m.
\end{align}
\textit{Sparse measure recovery} aims at recovering $\mu$ from the generalized moments $\Phi \mu$ \citep{de2012exact,duval2015exact}. This problem has been the focus of research in super-resolution, tensor decomposition, and the optimization of neural networks~\citep{chizat2021convergence}.

 Super-resolution relies on
 $\Omega = [0,1]$, and Fourier moments
 $\Phi(w)= [e^{-2\pi\Im  w}, \dots, e^{- 2\pi \Im m w}]$ where $\Im$ denotes the imaginary unit~\citep{candes2014towards}. Historically, super-resolution was proposed to enhance the resolution of optical devices in microscopy~\citep{mccutchen1967superresolution}, and medical imaging~\citep{greenspan2009super}. Various algorithms have been developed for super-resolution  \citep{moitra2015super,mccutchen1967superresolution,candes2014towards,bourguignon2007sparsity,baraniuk2010model,candes2014towards,tang2014near}. \citet{candes2014towards} propose a convex relaxation for super-resolution that only needs $m=\bigo(1/\ell)$ moments. This relaxation casts super-resolution to a semi-definite program(SDP) that can be solved in  $\bigo(\ell^{-\Delta -1}/\log(1/(\epsilon\ell)))$ time where $s^{-\Delta}$ is the time for matrix inversion of size $s$\footnote{$\Delta = 3$ for practical algorithms.}~\citep{lee2015faster}.

 For tensor decomposition, $\Omega$ is the unit sphere and  $\Phi(w):= \text{vector}(w^{\otimes k}) \in \R^{d^k}$. 
Tensor decomposition is information-theoretically possible only if the tensor degree is sufficiently large as $n<d^{k-1}$\citep{liu2001cramer}.  However, the complexity scales with the degree at an exponential rate. In that regard,  the best algorithm for tensor decomposition requires $\bigo(d^{\text{poly}(1/\epsilon)})$ time to reach an $\epsilon$-optimal solution~\citep{ma2016polynomial}. 

Neural networks use the following moments for the recovery
\begin{align*}
     \Phi(w) = \left[ \varphi(w^\top x_1), \dots, \varphi(w^\top x_m) \right] \in \R^m,  \quad \text{with} \; w \in \Omega =\R^d.
\end{align*}
The function $\varphi:\R \to \R$ is called an activation function, and $x_1, \dots, x_m$ are drawn i.i.d.~from an unknown distribution. The moments $\Phi \mu \in \R^m$ are the outputs associated with inputs $x_1,\dots, x_m$ for a single-layer network with weights $w_1,\dots, w_n$. This network is often called a planted (or teacher) network. A core topic in the theory of neural nets is the recovery of $\mu$ from  $\Phi\mu$ \citep{janzamin2015beating,bach2021gradient}. 

A line of research in physics studies \textit{cloning} a population of particles through interactions between the particles~\citep{carrillo2018measure}. The cloning is implemented by minimizing an energy function of probability measures. We will show that this optimization is an instance of measure recovery in the asymptotic regime of $m\to \infty$. In a mean-field regime when $n\to \infty$, an abstract infinite-dimensional optimization solves this specific instance of super-resolution~\citep{carrillo2018measure}. Our main contribution is the design of a polynomial-time algorithm for super-resolution based on this abstract optimization method.

% Non-convex sparse measure recovery is mainly studied in the regime $n\to \infty$, called the overparameterized regime \citep{bach2021gradient,carrillo2012mass}. In such asymptotic regime, \citep{bach2021gradient} proves the global convergence of gradient descent for neural nets, and \citep{carrillo2018measure} proves the global convergence of gradient descent on the energy distance for interacting particles. Implications of these theoretical findings on computational complexity have remained unknown.    

% \subsection*{Interacting particles}

% Interacting particles is a model to describe macroscopic behavior of physical systems using microscopic interactions between particles~\citep{friedli2017statistical}. For example, spine glasses are interacting particles with discrete states has extensively in physicists. Strikingly, physicists have achieved promising results for continuous-state interacting particles. We will derive a subtle connection between a particle system of interacting particles with spare measure recovery. 

\subsection{Main results}
The next theorem states our main contribution.

\begin{theorem}[An informal statement of Theorem~\ref{thm:finite_momments}]
    There exists an algorithm that obtains an $\epsilon$-solution for super-resolution in ${O(n^2 \ell^{-2}(1+(\ell/\epsilon)^2)}$ time using $m=O(n/\ell+n/\epsilon)$ moments. The algorithm  is designed based on a mean-field method analyzed by \citet{carrillo2018measure} in the regime $n,m\to \infty$. 
\end{theorem}
According to the last theorem, the proposed algorithm improves upon standard SDP solvers for super-resolution when $\ell =  o(n^{-2/(\Delta-1)})$ and $\epsilon = \Omega(n\ell^{0.5(\Delta+1)})$. For practical algorithms $\Delta =3$, hence $\ell=o(n^{-1})$ is a reasonable regime as $\ell\leq 1/n$. Furthermore, the proposed algorithm outperforms SDP solvers up to the $\Omega(n^{-1})$-accuracy.   This improvement demonstrates the power of mean-field optimization methods in algorithm design and calls for future studies in this line of research.

Our analysis leads to a result of independent interest: We establish the global convergence of gradient descent when optimizing the population loss for single-layer neural nets with weights on the upper-half unit circle, zero-one activations, and inputs drawn uniformly from the unit circle. To reach an $\epsilon$-accurate solution, GD needs only $\bigo(1/\epsilon)$ iterations. This result is inspired by theoretical studies of the mean-field regime $n\to \infty$.
    
   Finally, we extend our result for the recovery of sparse measures over $(d-1)$-dimensional unit sphere in polynomial time. 

\subsection{Limitations and discussions}

Our main focus is not to improve the state-of-art-method for the sparse measure recovery. But instead, we introduce a novel approach for algorithm design,  with potential broader applications. To further elaborate on this main message, we add remarks on the major limitations of our results.

The proposed algorithm requires $O(n/\ell)$ moments, while methods based on convex relaxation need only $O(1/\ell)$ moments. Furthermore, the established computational complexity is worse than SDP solvers for a small choice of $\epsilon$. While SDP solvers have been extensively studied and improved over the last century, our algorithm is the first practical algorithm developed by a mean-field optimization method. Compared to SDP solvers, the proposed algorithm is relatively simpler as it is an approximate gradient descent optimizing a non-convex function. It is promising that a simple (approximate) gradient descent enjoys a better computational complexity up to $\Omega(n^{-1})$-accuracy compared to efficient SDP solvers.  Future studies may improve the proposed algorithm or its analysis.

Our analysis for neural networks is for a specific example of two-dimensional neural networks and does not generalize to high-dimensional cases. Furthermore, our convergence result is established for the population training loss, not the empirical loss.  Despite these limitations, our result is a step towards bridging the gap between the mean-field and the practical regime of neural networks. 

To recover sparse measures over the high-dimensional unit sphere, we design specific moments. These moments are different from the moments used in super-resolution and neural networks. Yet, our result demonstrates the connection between the moments and the complexity of the recovery, which is under-studied in the literature. 

\subsection{Outline}
We bridge the gap between mean-field theory and the algorithm design for super-resolution in three steps: 
\begin{itemize}
    \item We first study super-resolution in the mean-field regime $n,m\to \infty$. In section~\ref{sec:mean_field}, we establish a subtle link between super-resolution and the optimization of the energy distance from physics.
    \item Leveraging the mean-field optimization of the energy distance, we prove a gradient descent method solves super-resolution in polynomial-time in $n$ when $m\to \infty$ as discussed in Section~\ref{sec:particle}.  
    \item Finally, we use a finite number of moments $m=O(n/\ell)$ to approximate gradient descent in Section~\ref{sec:poly}.
\end{itemize}
Sections~\ref{sec:nn}, and \ref{sec:hd} discuss the applications our analysis for specific instances of sparse measure recovery beyond super-resolution.
\section{A mean-field optimization for super-resolution} \label{sec:mean_field}

\subsection{Moment matching with the energy distance}

A standard approach for measure recovery is the moment matching method ~\citep{duval2015exact}. Assume that we are given Fourier moments of the measures $\nu$ and $\mu$ over $\R$, denoted by $f_\nu(t)$ and $f_\mu(t)$ respectively, where $f_\mu(t) = \int_{-\infty}^{\infty} e^{- 2\pi \Im t w}d\mu(w)$.
To match these moments, one can minimize the following function
\begin{align} \label{eq:energy_fourier}
    E(\nu) = \frac{1}{2\pi} \int_{-\infty}^{\infty}t^{-2}| f_\mu(t)-f_\nu(t)|^2 dt.
\end{align}
The above function is slightly different from standard mean squared loss for moment matching \citep{duval2015exact} as $t^{-2}$ enforces lower weights for high frequencies. Yet, the 
minimizer $\nu$ needs to obey $f_\nu(t) = f_\mu(t)$ for all $t \in \R$, similar to the standard square loss for the moment matching method~\citep{duval2015exact}.

A straightforward application of \textit{Parseval-Plancherel} equation yields the following alterinative expression for $E$ \citep{szekely2003statistics}:
\begin{multline} \label{eq:energy_dist}
    E(\nu) = 2  \int |w - w'| d\nu(w)d\mu(w') - \int | w - w'| d\nu(w) d\nu(w)  d\mu(w') - \int |w-w'| d\mu(w) d\mu(w').
\end{multline}
Indeed, $E$ is the energy distance between two probability measures $\nu$ and $\mu$. In physics, the energy distance is used to characterize the energy of interacting particles~\citep{di2013measure,carrillo2018measure}, where $\nu$ and $\mu$ represent populations of particles from two different species. 
The first term in $E$ describes interactions between the  populations. The second and third terms control internal interactions within each population. The energy distance admits the unique minimizer $\nu=\mu$~\citep{carrillo2018measure}. Therefore, minimizing $E$ clones the population $\mu$.

\subsection{Wasserstein gradient flows}
  The \textit{principle of minimum energy} postulates that the internal energy of a closed system (with a constant entropy) decreases over time~\citep{reichl1999modern}. Obeying this principle, the density of interacting particles evolves to reach the minimum energy. The density evolution can be interpreted as a gradient-based optimization \citep{ambrosio2005gradient}. Recall (proximal) gradient descent iteratively minimizes an objective function $f$ by the following iterative scheme 
 \begin{align*}
      w_{k+1} = \arg\min_{w\in \R^d} f(w) + \frac{1}{2 \gamma} \| w - w_k\|^2.
 \end{align*}
The above iterative scheme extends to general metric spaces, including the space of probability distributions with the Wasserstein metric denoted by $W_2$  \citep{ambrosio2005gradient}:
\begin{align*}
    \nu^{(k+1)}_\gamma =  \arg\min_{\nu}  E(\nu) + \frac{1}{2\gamma} W_2( \nu, \nu^{(k)})^2,
\end{align*}
where $\nu$ is a probability measure. Define $\nu_\gamma(t) = \{ \nu_{\gamma}^{(k)}, t \in (\gamma k,\gamma (k+1))\} $. As $\gamma\to 0$, $\nu_\gamma(t)$ converges to \textit{gradient flow} $\nu(t)$ obeying the continuity equation~\citep{jordan1998variational} as 
\begin{align*}
     \frac{d \nu}{dt} = \text{div}\left(\nu \left(\nabla \left(\frac{d E}{d\nu}\right)\right)\right),
\end{align*}
where div denotes the divergence operator. 
The above gradient flow globally optimizes various energy functions~\citep{bach2021gradient,mccann1997convexity,carrillo2018measure,carrillo2022global}. In particular, \citet{carrillo2018measure} have established the global convergence of the gradient flow for the energy distance. Yet, gradient flows are not always implementable.   

\section{Beyond mean-field optimization: Particle gradient descent} \label{sec:particle}
To numerically optimize the energy distance, one can restrict the support of the density to a finite set called particles, denoted by $\{v_1, \dots, v_n\}$, then minimize the energy w.r.t to the support~\citep{bach2021gradient,carrillo2022global}:
\begin{align*} 
 \min_{v_1, \dots, v_n} E\left( \frac{1}{n} \sum_{i=1}^n \delta_{v_i}\right).
\end{align*}
To optimize $E$, we may use first-order optimization methods such as subgradient descent: 
\begin{align} \tag{GD} \label{eq:GD}
    v_i^{(k+1)} = v_i^{(k)} - \gamma \frac{dE^{(k)}}{\|dE^{(k)}\|}, \quad  dE^{(k)}:= \frac{d E}{d v_i} \left( \frac{1}{n} \sum_{j=1}^n \delta_{v_j^{(k)}}\right) .
\end{align}
The above algorithm is called \emph{particle gradient descent} (GD)~\citep{chizat2022sparse,bach2021gradient}.
The normalization with the gradient norm is commonly used in non-smooth optimization~\citep{nesterov2003introductory}.
\citet{bach2021gradient} prove that the empirical distribution over $v_i^{(k)}$ converges to the Wasserstein gradient flow as $\gamma \to 0$ and $n\to \infty$ for a family of convex functions. Thus,  Wasserstein gradient flow is a mean-field model for particle gradient descent. It is not easy to analyze the convergence of particle gradient descent since $E$ is not a convex function. Taking inspiration from the convergence of Wasserstein gradient flow, the next lemma proves GD optimizes $E$   
 in polynomial-time.
 
 \begin{lemma} [GD for the energy distance] \label{lemma:particles}
 There exists a permutation of $[1, \dots, n]$ denoted by $\sigma$ such that
\begin{align*}
    \max_{i} |v_i^{(k)} - w_{\sigma(i)}| \leq \gamma
\end{align*}
holds
for $k\geq\floor{  \gamma^{-1} \max_{i} |v_i^{(0)} - w_{\sigma(i)}|}+1$ as long as $v_i^{(0)} \neq v_j^{(0)}$ for all $i\neq j$.
\end{lemma}

\begin{figure}[t!]
    \centering
    \includegraphics[width=0.5\textwidth]{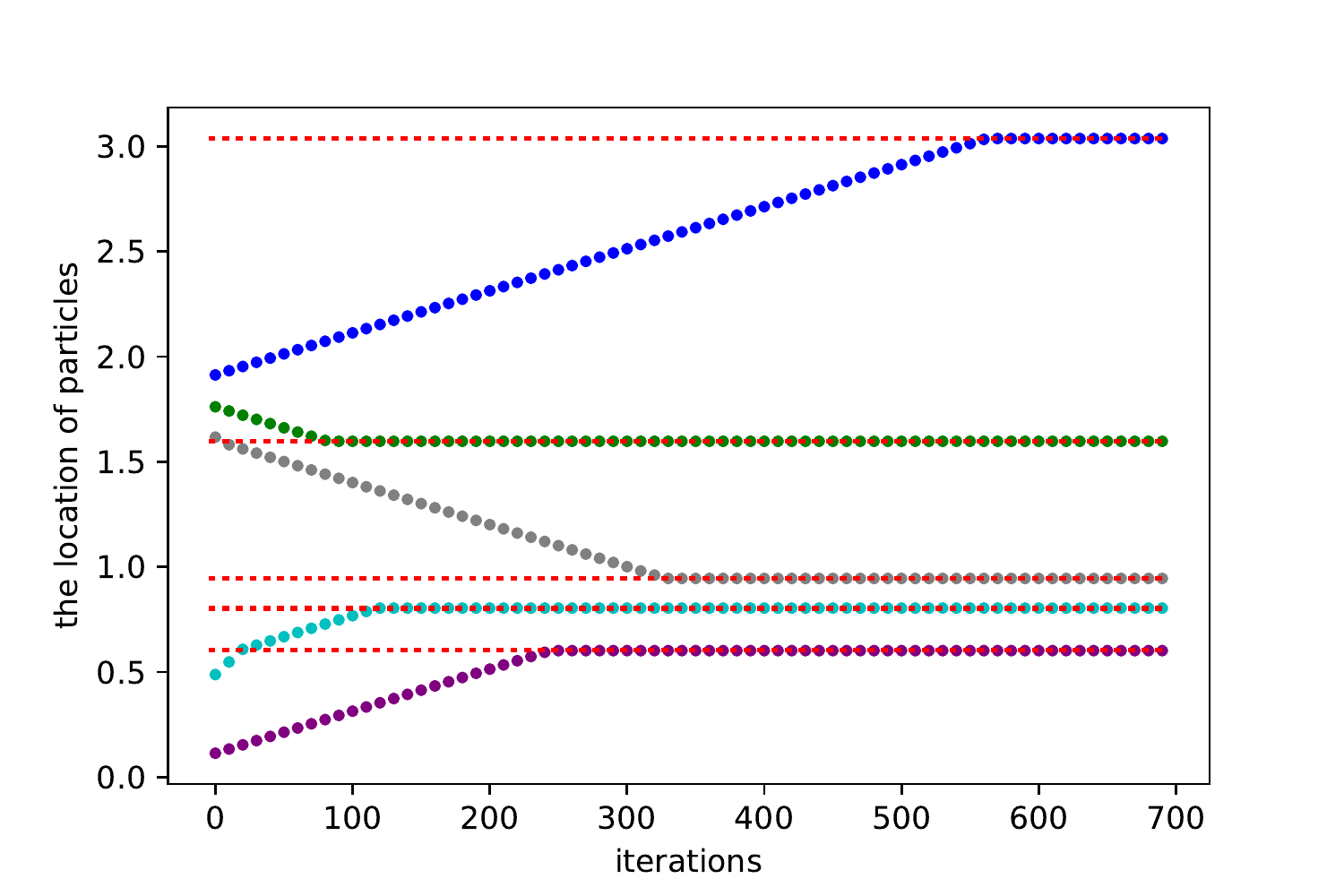}
    \caption{\textbf{The convergence of particle gradient descent on the energy distance.} The red dashed lines mark $w_1,\dots, w_5$ which are  drawn i.i.d. from uniform$[0,\pi]$. The other curves trace  $v_{\text{blue}}^{(k)}, \dots, v_{\text{purple}}^{(k)}$ (on the vertical axis) for $k$ shown on the horizontal axis.   The stepsize $\gamma$ is set to $0.01$ for this experiment.}
    \label{fig:gd}
\end{figure}
Informally speaking, the particles from two species contract with decreasing stepsize. With stepsize $\gamma=\epsilon$,  GD yields an $\epsilon$-solution in $\bigo(1/\epsilon)$. Figure~\ref{fig:gd} illustrates a numerical simulation of GD where we observe the convergence of particles with iterations. Interestingly, $v_i^{(k)}$ converges to $w_i$ when $v_1^{(0)}\leq \dots \leq v_n^{(0)}$ and $w_1\leq \dots \leq w_n$. Such convergence is due to the monotone property of optimal transport leveraged in the mean-field analysis of the Wasserstein gradient flow~\citep{carrillo2018measure}. We also use this convergence property to prove the last lemma. Furthermore, we use
the sparsity of $\mu$ for the proof. While the sparsity is not met in the context of Wasserstein gradient flows, often emerges in signal processing and super-resolution.

As discussed above, the Wasserstein gradient flow on the energy distance solves super-resolution in the mean-field regime $n,m\to \infty$. To design an implementable algorithm, we need to use a finite $n$ and $m$. According to the last lemma, particle gradient descent on the energy distance has a polynomial complexity in $n$, but not in $m$. Since super-resolution is limited to a finite $m$, we design an algorithm with a polynomial complexity in $m$ and $n$ when $m=\bigo(n/\ell)$ in the next section.

% The proof of the last Theorem is based by establishing a Lyapunov stability for gradient descent. In other words, we prove that there exists a function 
% Indeed, we first skip the sensing mechanism; instead, we focus on the recovery of a sparse measure by optimizing the energy distance. First, we prove that subgradient descent optimizes $E$; then, we prove it is possible to implement subgradient descent on $E$ with 
\noindent
% \textbf{Background.}
%  Population $\nu$ evolves till it achieve the minimal energy. This evolution is described by a specific Partial Differential Equation (PDE): 

% \begin{itemize}
%     \item PDE
%     \item Gradient flow 
%     \item Convergence result 
% \end{itemize}

% \textbf{Particle gradient descent.}
% \begin{itemize}
%     \item Introduce particle gradient descent
%     \item Results for sparse $\mu$
%     \item Results for non-sparse $\mu$
%     \item Remarks on applications Make the transition to the sparse deconvolution and sparse measure recovery
% \end{itemize}

% \textbf{Convergence of gradient descent.}
% $E$ is not convex in $\nu$ or $\mu$. Surprisingly, \citep{di2013measure} proves that gradient flow in $\nu$ and $\mu$ is global convergent. Gradient flow with respect to probability measure is a notion in optimal transport that generalizes the notion of optimization with gradient descent to probability measures. Implementing gradient flow is not computationally feasible since $\mu$ and $\nu$ are infinite dimensional measures. Here, we prove that when $\nu$ and $\mu$ are sparse (obeying \ref{}), then it is possible to optimize $H$ with respect to $\nu$. 

\section{A novel polynomial-time algorithm for super-resolution} \label{sec:poly}

We approximate GD using a finite number of Fourier moments. The proposed approximation is inspired by Fourier features~\citep{rahimi2007random} used to efficiently approximate positive-definite kernels. Using a similar technique, we propose an approximate GD that provably solves super-resolution.

The subgradient of the energy distance has the form 
\begin{align}
    n^2\frac{d E}{ d v_i } = \sum_{j=1}^n 2 \left( \sign(v_i - w_j) - \sign(v_i - v_j)\right) .
\end{align}
An application of the Fourier series (on a bounded interval) allows us to write the $\sign$ term in the above equation as
\begin{align*}
    \sign(v - w) = -\Im \sum_{i=1}^{\infty} c_i e^{-\Im 2\pi i (v-w)}, \quad c_{i} = \begin{cases}
        \frac{4}{\pi i} & \text{odd } i \\ 
        0 & \text{even } i
    \end{cases},
\end{align*}
where $w \neq v$.
Since the $c_i$s are non-negative, we can alternatively write the $\sign$ function as
\begin{align*}
     \sign( v - w) & = -\Im \sum_{i=1}^\infty  \left(\sqrt{c_i} e^{-\Im 2\pi i v}\right) \left(\sqrt{c_i} e^{ \Im 2\pi  i w}\right).
\end{align*}
   Cutting-off high frequencies leads to an approximation for the $\sign$: 
 \begin{align} \label{eq:approx}
     \sign(v-w) \approx -\Im \langle \sqrt{c} \odot \Phi^* (w),\sqrt{c} \odot \Phi (v) \rangle,\quad \Phi(w) = [e^{-\Im 2\pi w}, \dots, e^{-\Im 2\pi m w}],
 \end{align}
 where $*$ indicates the conjugate of a complex number and $c := [c_1, \dots, c_m]$. 
Using the above approximation, we propose an approximate GD in Algorithm~\ref{alg:recovery} for super-resolution.
\begin{algorithm}
\caption{Polynomial-time algorithm for super-resolution}\label{alg:recovery}
\begin{algorithmic}
\Require $\Phi\mu$, and stepsize $\gamma$
\vspace{0.2cm}
\State Recall $\Phi(w) = [e^{-\Im 2\pi w}, \dots, e^{-\Im 2\pi m w}]$.
\State Let $c = [c_1, \dots, c_m]$ where $c_i$ s denote the coefficients of the Fourier series expansion for the $\sign$ on $[-\pi, \pi]$
\State Let $k = \floor{ 200\pi\gamma^{-1}}+1$.

\For{$q = 1,\dots, k$}
\For{$i=1,\dots,n$}
 \State Let $\nu^{(q)} = \frac{1}{n} \sum_{j=1}^n \delta_{v_j^{(q)}}$ 

\State $ d \widehat{E} := -\Im \langle \sqrt{c} \odot (\Phi \mu)^*,\sqrt{c} \odot \Phi \nu^{(q)} \rangle -\sum_{j=1}^n \sign\left(v_i^{(q)} - v_j^{(q)}\right)$

\State $v_{i}^{(q+1)} = v_{i}^{(q)} -\gamma  \frac{d \widehat{E}}{\left\|d \widehat{E}\right\|} $ \;
\EndFor
\EndFor
\State \Return $v_1^{(k)}, \dots, v_n^{(k)}$\;
\end{algorithmic}
\end{algorithm}

\begin{figure}[t!]
    \centering
\includegraphics[width=0.5\textwidth]{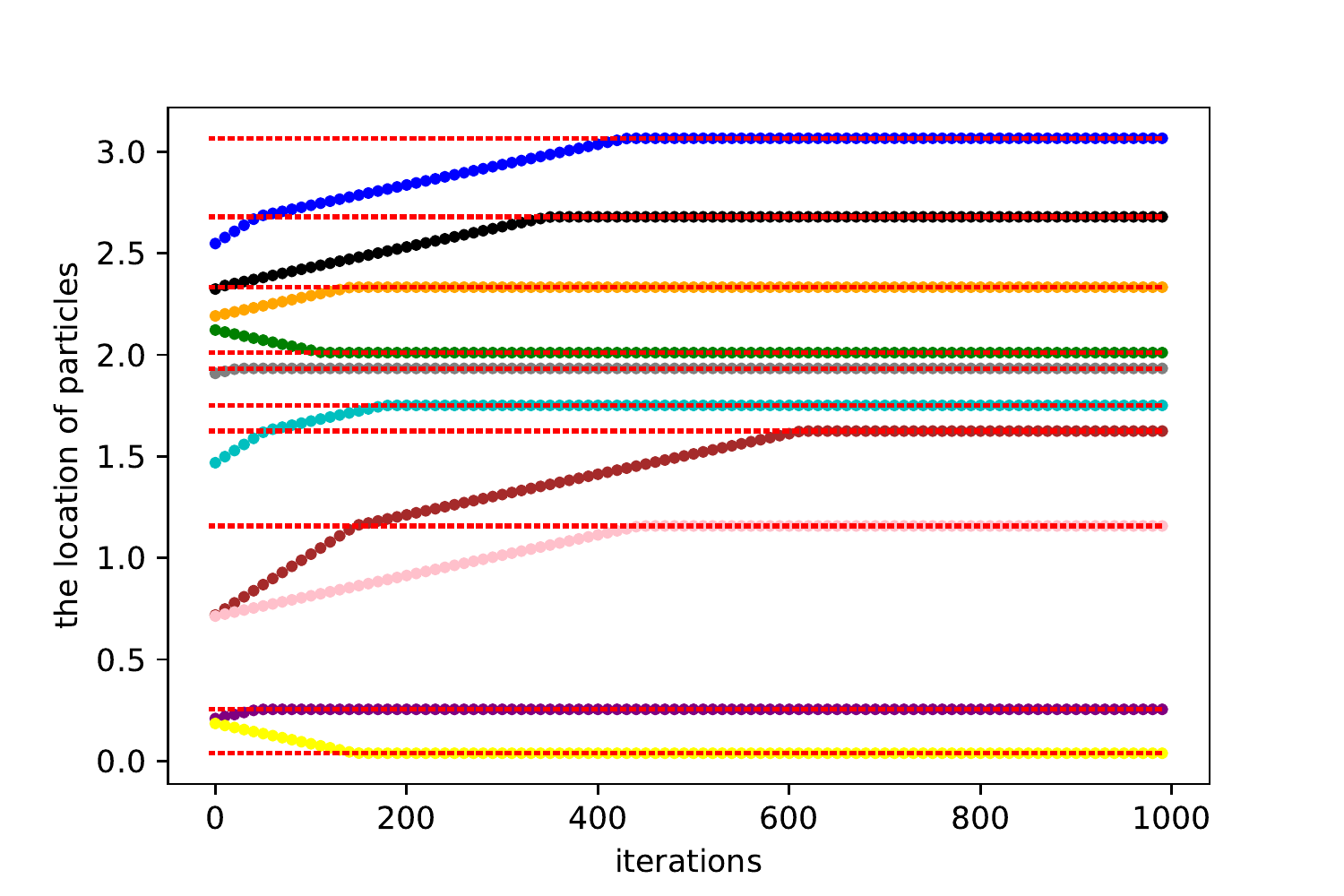}
    \caption{\textbf{A simulation for Algorithm~\ref{alg:recovery}.} The red dashed lines show the location of $w_1,\dots, w_{10}$ which are  drawn i.i.d. from uniform$[0,\pi]$. The other lines trace the dynamic of $v_{\text{blue}}^{(k)}, \dots, v_{\text{yellow}}^{(k)}$ (the vertical axis) with $k$ (the horizontal axis).   We use $\gamma=0.01$ and $m=200$ in this experiment. The convergence to the red dashed lines confirms Algorithm~\ref{alg:recovery} solves this random instance of super-resolution.}
    \label{fig:approximate_gd}
\end{figure}

\begin{theorem} \label{thm:finite_momments}
Assume \ref{assume:distinct}($\ell$) holds, and $w_i \in \Omega \subset [0, \pi]$ for all $i \in \{1, \dots, n\}$. Suppose that
Algorithm~\ref{alg:recovery} starts from $\{v_{i}^{(0)} \in [0, \pi]\}_{i=1}^n$ where $v_{i}^{(0)}\neq v_j^{(0)}$ for $i\neq j$, and returns $v_1, \dots, v_n$; then, there exists a permutation of $\{1, \dots, n\}$ denoted by $\sigma$ such that
\begin{align}
   \min_i |v_i - w_{\sigma(i)}|   \leq \epsilon
\end{align}
holds for $
   \gamma = \min\left\{ \epsilon/3,\ell \right\}$, and $m= \Omega(n\epsilon^{-1} + n\ell^{-1})$ in $\bigo(n^2/\epsilon^2+ n^2/\ell^2)$ time.
\end{theorem}
Thus, Algorithm~\ref{alg:recovery} solves super-resolution in polynomial-time. The proof of the above Theorem, presented in the Appendix, is based on the global convergence of GD on the non-convex energy distance established in Lemma~\ref{lemma:particles}. Figure~\ref{fig:approximate_gd} shows the dynamics of Algorithm~\ref{alg:recovery} in a numerical simulation.

Although the measure $\mu$ is defined over $\R$, the recovery of $n$-sparse measures over $\R$ is an $n$-dimensional optimization problem. Searching over $n$-dimensional parameters suffers from  $O((1/\epsilon)^n)$ computational complexity \citep{nesterov2003introductory}. According to the last theorem, the proposed algorithm significantly improves upon brute-force search. To further elaborate on the efficiency of Algorithm~\ref{alg:recovery}, we compare its performance to an efficient method based on convex relaxation.

The sparsity is a non-convex constraint on measures. The total variation of the measure $\mu$, denoted by $\|u \|_{\text{TV}}$, is a convex function that relaxes this constraint. Leveraging the variation norm, \citet{candes2014towards} introduce the following convex program for super-resolution
\begin{align} \label{eq:convex_relax} \tag{convex relaxation}
     \min_{\nu} \lambda \| \nu \|_{TV} + \|  \Phi_{\mu} - \Phi_\nu\|^2, 
\end{align}
where $\nu$ is a probability measure over $\Omega$.
The unique solution of the above program is $\mu$ under \ref{assume:distinct}($\ell$) as long as $m=\Omega(1/\ell)$ \citep{candes2014towards}. Yet, the above program is infinite-dimensional, hence challenging to solve. One way to approximate the solution of the above problem is using finite support for $\nu$. However, a fine grid is needed for an accurate recovery \citep{candes2014towards,tang2014near}. A fine grid not only causes numerical instabilities but also violates the coherence condition required for the recovery~\citep{tang2014near}.  \citet{candes2014towards} propose a different method to solve the above program. 
The dual of the \ref{eq:convex_relax} is a finite-dimensional semidefinite program. This program has a conic constraint on a $m\times m$ matrix with $m$ constraints, hence efficient cutting plane methods need at least $O(\ell^{-\Delta-1}\log(1/\epsilon))$ time to solve this problem~\citep{lee2015faster} where $m^{\Delta}$ is the complexity of matrix inversion for matrices of size $m$ (for practical algorithms, $\Delta =3$). In comparison, Algorithm~\ref{alg:recovery}  requires $O(n^2 \ell^{-2}\left(1+\frac{\ell}{\epsilon}\right)^2)$ time which improves upon $\ell$, while suffers from a worse complexity in $\ell/\epsilon$ and $n$. As long as $\ell =  o(n^{-2/(\Delta-1)})$ and $\epsilon = \Omega(n\ell^{0.5(\Delta+1)})$, the proposed algorithm improves upon the state of the art. Interestingly, this improvement is achieved by non-convex optimization.

The conventional Prony’s and matrix pencil methods \citep{moitra2015super} can not be used for the specific settings of super-resolution considered in this paper. For example, the matrix pencil method needs individual measurements $\exp(-\Im 2\pi \omega_k)$. But, we only use the average of measurements for the recovery akin to \citet{candes2014towards}.

Algorithm~\ref{alg:recovery} is inherently robust against noise. This algorithm is an approximate  GD on the energy distance. To prove Theorem~\ref{thm:finite_momments}, we show that GD tailors $O(1/n)$ approximate gradient to recover an approximate solution. Although we analyzed the specific noise imposed by cutting off Fourier moments, it is easy to prove that an additional zero-mean noise with variance $o(1/n)$ changes the bounds only up to constants in expectation. 

\section{Global optimization for a toy neural network} \label{sec:nn}
While the optimization of neural nets is NP-hard in the worst case~\citep{blum1992training},  GD obtains almost the minimum error in practice for various neural nets~\citep{he2016deep}. A line of research attributes this surprising performance to overparameterization. For neural networks with a single hidden layer,  \citet{bach2021gradient} study the evolution of the density of neuron weights obeying GD. As the number of neurons tends to infinity, the density dynamics make a gradient flow that globally optimizes the training loss of neural nets \citep{bach2021gradient}. 
 
 Taking inspiration from mean-field analyses, we establish the global convergence of gradient descent for a specific toy neural network. In particular, we show that the training of a specific neural network can be cast as the minimization of the energy distance. Suppose $\widehat{w}_1, \dots, \widehat{w}_n$ are on the upper-half unit circle, hence admit polar representations $\widehat{w}_i = (\sin(w_i),\cos(w_i))$ where $w_1,\dots, w_n \in [0,\pi]$. Using these vectors, we construct a teacher neural network as
 \begin{align}
      f(x) := \frac{1}{n}\sum_{i=1}^n \varphi(x^\top \widehat{w}_i), \quad \quad \varphi(a) = \begin{cases} 
     1 & a>0 \\
     0 & a \leq 0
     \end{cases}, \quad x \in \R^2.
 \end{align}
 Note that $\widehat{w}_i$ are the weights of the neurons. The function $\varphi$ is the zero-one activation, which is used in the original model of McCulloch-Pitts for biological neurons~\citep{jain1996artificial}. An important problem in the theory of neural networks is the recovery of $\widehat{w}_i$  \citep{janzamin2015beating} from $f(x)$ when $x$ is drawn from a distribution. \citet{janzamin2015beating} elegantly reduce this problem to tensor decomposition. However, methods for tensor decomposition suffer from an exponential rate with the approximation factor: $O(2^{\text{poly}(1/\epsilon)})$ for an $\epsilon$-solution~\citep{ma2016polynomial}. To the best of our knowledge, there is no algorithm with a polynomial complexity in $\epsilon$ and $n$.  
 
 In practice, tensor decomposition is not used to optimize neural networks. Often neural nets are optimized using simple GD optimizing a non-convex function. Let $\widehat{v}_1, \dots, \widehat{v}_n \in \R^2$ be the weights of a neural network with output function $\frac{1}{n}\sum_{i=1}^n \varphi(x^\top \widehat{v}_i)$. Neural networks may optimize the mean square error:
 \begin{align}
      L\left(\widehat{v}_1,\dots, \widehat{v}_n \right) =  \E_{x} \left(\frac{1}{n}\sum_{i=1}^n \varphi(x^\top \widehat{v}_i) - \frac{1}{n}\sum_{i=1}^n \varphi(x^\top \widehat{w}_i) \right)^2.
 \end{align}
  The next proposition states the above objective is equivalent to the energy distance when the input $x$ is drawn uniformly from the unit circle. 
 
 \begin{Proposition} \label{lemma:nn}
 For $x$ is drawn uniformly from the unit circle,
 \begin{align}
     (\pi n^2) L\left(\{\widehat{v}_i=(\sin(v_i),\cos(v_i))\} \right) = E\left(\frac{1}{n}\sum_{i=1}^n \delta_{v_i}\right)
 \end{align}
 holds, where $E$ is the energy distance defined in Eq.~\eqref{eq:energy_dist}.
 \end{Proposition}
 Thus, neurons can be interpreted as interacting particles. Combining this with Lemma~\ref{lemma:particles} concludes the global convergence of GD on $L$.

\section{ High dimensional sparse measure recovery} \label{sec:hd}

To the best of our knowledge, there is no polynomial-time algorithm for the recovery of sparse measures over $\R^d$. As noted before, tensor decomposition has an exponential time complexity in the approximation factor. \citet{bredies2013inverse} develop a conditional gradient method that obtains an $\bigo(\text{iterations}^{-1})$-optimal solution. Yet,  each iteration of the conditional gradient method is a non-convex optimization problem whose complexity has remained unknown. In this section, we demonstrate our analysis can provide insights into the complexity of the recovery of measures over $\R^d$.

Recall various instances of sparse measure recovery leverage different types of moments: polynomial moments for tensor decomposition, non-polynomial ridge-type moments for neural networks, and Fourier moments for super-resolution. This variety of moments arise an important question: Is there any specific moment type that allows the recovery in polynomial time? Using Theorem~\ref{thm:finite_momments}, we prove that there are moments that allow a polynomial-time recovery. 
\begin{lemma} \label{lemma:high_dim}
    Suppose that measure $\mu$ has the support of size $n$ over unit $(d-1)$-dimensional unit sphere. There exist specific moments that allow us to recover $\mu$ from the moments with Algorithm~\ref{alg:recovery} in $\bigo\left( (nd)^2 \epsilon^{-2} + (nd)^4 \ell^{-2} \right)$ time with high probability. 
\end{lemma}
To prove the above theorem, we reduce a $d$-dimensional recovery to a set of one-dimensional instances and run Algorithm~\ref{alg:recovery} on each of them. The formal statement and detailed proofs are outlined in the Appendix. 

The last lemma highlights the important role of the moments in the complexity of sparse measure recovery, which is under-studied in the literature. For example, the recovery from the specific moments used in neural networks may be easier compared to tensor decomposition which potentially explains the power of neural networks. The last lemma motivates future research in this vein.    
\section{Conclusion}
We prove that mean field theory has applications beyond theoretical analyses. For the case study of super-resolution, we develop a polynomial time algorithm from an abstract optimization, which operates in the theoretical mean-field regime of infinite problem size. To highlight the power of such algorithm design, we prove that the proposed algorithm improves upon standard methods in a particular parameter regime.  Furthermore, we demonstrate the application of our analysis in the broader context of sparse measure recovery. By intersecting theoretical physics with computer science, our findings call for future research in algorithm design based on mean-field optimization.
\section*{Acknowledgments and Disclosure of Funding}
We thank Lenaic Chizat for his helpful discussions. Hadi Daneshmand received funding from the Swiss Na-
tional Science Foundation for this project (grant P2BSP3 195698). We also acknowledge
support from the European Research Council (grant SEQUOIA 724063) and the French government under management of Agence Nationale de la Recherche as part of the “Investisse-
ments d’avenir” program, reference ANR-19-P3IA-0001(PRAIRIE 3IA Institute).

\bibliographystyle{apalike}  

\bibliography{refs}
\appendix
\section{Interacting particles}
This section proves Lemma~\ref{lemma:particles}. The proof uses two important tools from the physics of interacting particles: \begin{itemize}
    \item[(i)] Akin to \citet{carrillo2018measure}, our proof is based a Lyapunov analysis. We will show a Wasserstein metric decays with GD iterations.
    \item[(ii)] We use the monotonicity of the optimal transport map established by \citet{carrillo2012mass}. 
\end{itemize} 
\subsection{Optimal transport and sorting} \label{sec:distance}
Let $\nu = \frac{1}{n} \sum_{i=1}^n \delta_{v_i}$, and $\mu = \frac{1}{n} \sum_{i=1}^n \delta_{w_i}$.
We introduce the following distance notion
\begin{align}
     W(\nu,\mu) = \min_{\sigma \in \Lambda}\max_{i} |v_{\sigma(i)} - w_{i}|,
\end{align}
where $\Lambda$ is the set of all permutations of indices $\{1, \dots, n\}$. Indeed, $W$ is Wasserstein-$\infty$ distance and the permutation $\sigma$ is a transport map from $\mu$ to $\nu$~\citep{santambrogio2015optimal}. The optimal transport map minimizes $\max_{i} |v_{\sigma(i)} - w_{i}|$.  The next lemma proves that a permutation $\sigma$ sorting $v_{\sigma(i)}$ is the optimal transport map. Notably, this lemma is an extension of  Lemma 1.4 of \citet{carrillo2012mass}. 
\begin{lemma}[Optimal transport] \label{lemma:optimal_transport}
  For $v_{\sigma(1)} \leq \dots \leq v_{\sigma(n)}$, the following holds:
 \begin{align}
      W(\nu, \mu) = \max_{i \in \{1,\dots, n\}}| v_{\sigma(i)}-w_i |.
 \end{align}
\end{lemma}
\begin{proof}
 Let $\sigma^*$ denote the optimal transport obeying
\begin{align}
     \sigma^* = \arg_{\sigma \in \Lambda}\underbrace{\min_i |v_{\sigma(i)} - w_i|}_{f(\sigma)}.
\end{align}
The proof idea is simple: if there exists $i<j$ such that $v_{\sigma^*(i)}>v_{\sigma^*(j)}$, then swapping  $\sigma^*(i)$ with $\sigma^*(j)$ will not increase $f$. To formally prove this statement, we define the permutation $\sigma'$ obtained by swapping indices in $\sigma^*$ as
\begin{align}
     \sigma'(q) = \begin{cases} 
     \sigma^*(q) & q\neq i \\ 
     \sigma^*(j) & q = i \\
     \sigma^*(i) & q=j.
     \end{cases}
\end{align}
We prove that $f(\sigma') \leq f(\sigma)$.
Let define the following compact notations:
\begin{align}
    \Delta_{ij} & = \max \{ | v_{\sigma^*(i)} - w_i |, | v_{\sigma^*(j)} - w_j |\}, \\
    \Delta'_{ij} & = \max \{ | v_{\sigma'(i)} - w_i |, | v_{\sigma'(j)} - w_j |\} = \max \{ | v_{\sigma^*(j)} - w_i |, | v_{\sigma^*(i)} - w_j |\} .
\end{align}
\noindent According to the definition, 
\begin{align}
     f(\sigma') = \max \{ \Delta'_{ij}, \max_{q\neq i,j} | v_{\sigma^*(q)} - w_q | \}\leq  \max \{ \Delta_{ij}, \max_{q\neq i,j} | v_{\sigma^*(q)} - w_q | \} =f(\sigma^*)
\end{align}
holds.
Since $v_{\sigma^*(i)}>v_{\sigma^*(j)}$, $\Delta'_{ij}< \Delta_{ij}$ holds as  it is illustrated in the following figure. 
\tikzset{
  mynode/.style={fill,circle,inner sep=2pt,outer sep=0pt}
}
\begin{figure}[h!]
    \centering
    \begin{tikzpicture}
    \draw[olive,thick,latex-latex] (0,0) -- (7,0);
    \node[mynode,fill=red,label=above:\textcolor{red}{$w_i$}] (wi) at (2,0)  {};
    \node[mynode,fill=red,label=above:\textcolor{red}{$w_j$}] (wj) at (3,0)  {};
     \node[mynode,fill=blue,label=above:\textcolor{blue}{$v_{\sigma^*(j)}$}] (thj) at (5,0)  {};
      \node[mynode,fill=blue,label=above:\textcolor{blue}{$v_{\sigma^*(i)}$}] (thi) at (6,0)  {};
%     node[pos=0.4,mynode,fill=red,text=blue,label=above:\textcolor{red}{$w_j$}]{}
%     node[pos=0.6,mynode,fill=blue,text=green,label=above:\textcolor{blue}{$v_{\pi^*(j)}$}]{}
%     node[pos=0.8,mynode,fill=blue,text=green,label=above:\textcolor{blue}{$v_{\pi^*(i)}$}]{};
\draw [
    thick,
    decoration={
        brace,
        mirror,
        raise=0.5cm
    },
    decorate
] (wi) -- (thi) node [pos=0.5,anchor=north,yshift=-0.55cm] {$\Delta_{ij}$};
\draw [
    thick,
    decoration={
        brace,
        raise=1cm
    },
    decorate
] (wi) -- (thj) node [pos=0.5,anchor=south,yshift=1cm] {$\Delta'_{ij}$};
  \end{tikzpicture}
\end{figure}

\noindent Therefore, $f(\sigma')  \leq W(\nu,\mu)$ holds, in that $\sigma'$ is also an optimal transport. Replacing $\sigma^*$ by $\sigma'$ and repeating the same argument inductively concludes the proof. 
\end{proof}

\paragraph{The gradient direction.}
 The gradient of $E$ can be expressed by cumulative densities as 
\begin{align*}
    \frac{d E}{dv_i} = 2 \left( \int^{v_i}_{-\pi} \mu(v) dv- \int^{v_i}_{-\pi} \nu(v) dv\right), \quad \nu = \frac{1}{n}\sum_{i=1}^n \delta_{v_i}.
\end{align*}
Thus, moving $v_i$ along the negative gradient compensates the difference between the cumulative densities of $\mu$, and $\nu$. The next lemma formalizes this observation.
\begin{lemma}[Gradient direction] \label{lemma:gradient_dir}
Let $\sigma^*$ is the optimal transport from $\mu$ to $\nu$, then following bound holds
 \begin{align} \label{eq:decay_condition}
  \sign\left(v_{\sigma^*(i)} - w_{i}\right) \left( \frac{d E}{d v_{\sigma^*(i)}} \right) \geq 1.
\end{align}
\end{lemma}
 
\begin{proof}[Proof of Lemma~\ref{lemma:gradient_dir}]
We prove the statement for $v_1\leq \dots \leq v_n$. For the general proof,  replace $v_i$ by $v_{\sigma^*(i)}$.
The partial derivative $dE/dv_i$ consists of two additive components:
 \begin{align}
       \frac{d E}{d v_i }  = \underbrace{\sum_{j} \text{sign}(v_i - w_j)}_{\Delta} -\underbrace{\sum_{j\neq i} \text{sign}(v_i - v_j)}_{2i-n-1}, \label{eq:grad_structure}
    \end{align}
where 
\begin{align}
            \Delta & =  \left|\{ w_m < v_i \}\right|-|\{w_m > v_i  \}| \\ 
            & = 2 \left|\{ w_m < v_i \}\right| - n  \label{eq:delta_decompostion1}\\ 
            & = n - 2 |\{w_m > v_i  \}|. \label{eq:delta_decompostion2}
\end{align}
Consider the following two cases:
   \begin{itemize}
       \item[i.] $v_i>w_i$: 
       In this case, $|\{ w_m<v_i\}|\geq i$ (see Fig.~\ref{fig:elements_bound}). Plugging this into Eq.~\eqref{eq:delta_decompostion1}, we get $\Delta \geq 2i-n$ that yields $d E / d v_i \geq 1/(\pi n^2)$.
       \item[ii.] $v_i<w_i$: In this case, $| \{ w_m \geq v_i \} | \geq n-i+1$ demonstrated in Fig.~\ref{fig:elements_bound}. Using Eq.~\eqref{eq:delta_decompostion2}, we get  $\Delta \leq 2i-n-2$ that leads to $d E / d v_i \leq -1/(\pi n^2) $. 
   \end{itemize}
   Combining the above two results concludes the proof. 
   \tikzset{
  mynode/.style={fill,circle,inner sep=2pt,outer sep=0pt}
}
   \begin{figure}[h!]
    \centering
    \begin{tabular}{c c}
     \begin{tikzpicture}
    \draw[olive,thick,latex-latex] (0,0) -- (7,0);
     \node[mynode,fill=red,label=above:\textcolor{red}{$w_1$}] (w1) at (1.5,0)  {};
     \node[label=above:\textcolor{red}{$\dots$}] (wk) at (2.25,0)  {};
    \node[mynode,fill=red,label=above:\textcolor{red}{$w_m$}] (wi) at (3,0)  {};
      \node[mynode,fill=blue,label=above:\textcolor{blue}{$v_{i}$}] (thi) at (4,0)  {};

\draw [
    thick,
    decoration={
        brace,
        mirror,
        raise=0.5cm
    },
    decorate
] (w1) -- (wi) node [pos=0.5,anchor=north,yshift=-0.55cm] {$|\{ w_m < v_i \}| \geq i $};

  \end{tikzpicture} &  \begin{tikzpicture}
    \draw[olive,thick,latex-latex] (0,0) -- (7,0);
    \node[mynode,fill=blue,label=above:\textcolor{blue}{$v_i$}] (thi) at (2,0)  {};
    \node[mynode,fill=red,label=above:\textcolor{red}{$w_m$}] (wi) at (3,0)  {};
    \node[label=above:\textcolor{red}{$\dots$}] (wm) at (4,0)  {};
     \node[mynode,fill=red,label=above:\textcolor{red}{$w_n$}] (wn) at (5,0)  {};

\draw [
    thick,
    decoration={
        brace,
        mirror,
        raise=0.5cm
    },
    decorate
] (wi) -- (wn) node [pos=0.5,anchor=north,yshift=-0.55cm] {$| \{ w_m > v_i \} | \geq n-i+1$};

  \end{tikzpicture} 
    \end{tabular}
    \caption{The cardinality bound. Left: $v_i>w_i$. Right: $v_i<w_i$.}
    \label{fig:elements_bound}
\end{figure}
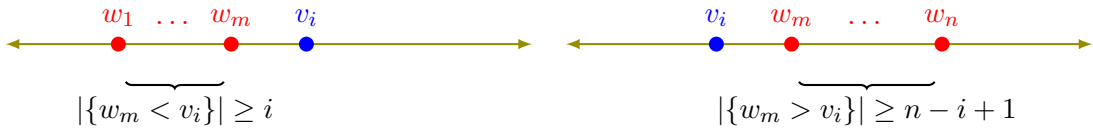
\end{proof}
\subsection{Subgradient descent}
Recall the recurrence of subgradient descent as 

\begin{align}
    v_i^{(k+1)} = v_i^{(k)} - \gamma \frac{dE^{(k)}}{|dE^{(k)}|}, \quad  dE^{(k)}:= \frac{d E}{d v_i} \left( \nu^{(k)}\right), \quad \nu^{(k)} = \frac{1}{n} \sum_{i=1}^n \delta_{v_i^{(k)}}  .
\end{align}
\noindent According to Lemma~\ref{lemma:gradient_dir}, gradient descent contracts $v_i$ to $w_i$ for sorted $v_i$s and $w_i$s, which is the optimal transport map. Combining these two observations, the next lemma establishes the convergence of gradient descent in terms of $W$ distance. 
\subsection{Proof of Lemma~\ref{lemma:particles}}

Let $\sigma^*$ is the optimal transport map from $\nu^{(k)}$ to $\mu$. 
  \begin{align}
      \left| v_{\sigma^*(i)}^{(k+1)} - w_i \right|= \left|v_{\sigma^*(i)}^{(k)} - w_i  - \gamma \frac{dE^{(k)}}{|dE^{(k)}|} \right|.
  \end{align}
  Invoking Lemma~\ref{lemma:gradient_dir} yields
  \begin{align}
       \left| v_{\sigma^*(i)}^{(k+1)} - w_i \right| & = \left|v_{\sigma^*(i)}^{(k)} - w_i  - \gamma \frac{dE^{(k)}}{|dE^{(k)}|}\right| \\
       & = \left|v_{\sigma^*(i)}^{(k)} - w_i  - \gamma\sign(v_{\sigma^*(i)}^{(k)} - w_i )\right|.
  \end{align}
 Using Lemma~\ref{lemma:optimal_transport}, we get 
 \begin{align}
     W(\nu^{(k+1)},\mu) & \leq  \max_{i} \left| v_{\sigma^*(i)}^{(k+1)} - w_i \right| \\ 
     & \leq \max_i \left|\left|v_{\sigma^*(i)}^{(k)} - w_i\right|  - \gamma \right| \\ 
     & \leq \max\{ W(\nu^{(k)},\mu)-\gamma ,\gamma\}  \label{eq:convergencebound_gd}.
 \end{align}
 
 We complete the proof by contradiction. Suppose that $W(\nu^{(q)},\mu) \geq \gamma$ holds for $q=1,\dots, q'$ where $q' = \floor{  W(\nu^{(0)},\mu)/\gamma}+1$; then, the induction over $k$ yields
 \begin{align}
       W\left(\nu^{(q')},\mu\right) < W(\nu^{(0)},\mu) - W(\nu^{(0)},\mu) <0.
 \end{align}
 The above inequality contradicts to $W\geq 0$. Therefore, there exists $q_*<q'$ such that ${W(\nu^{(q_*)},\mu) \leq \gamma}$. According to Eq.~\eqref{eq:convergencebound_gd}, $W(\nu^{(q)},\mu) \leq \gamma$ holds for all $q\geq q_*$.

\section{Super resolution} \label{sec:finite_moments}
In this section, we prove Theorem~\ref{thm:finite_momments}.
Recall, $\Phi$ are obtained by the Fourier series expansion of the $\sign$ on the interval $[-\pi, \pi]$ as
\begin{align}
    \Delta \neq 0 \text{ and } \Delta \in (-\pi, \pi): \sign(\Delta)  = \sum_{r=0}^\infty \left(\frac{4}{\pi (2r+1)}\right) \sin((2r+1)\Delta).
\end{align}
The Fourier series expansion of order $2m+1$ is denoted by $g$ as
\begin{align}
    g(\Delta) =  \sum_{r=0}^m \left(\frac{4}{\pi (2r+1)}\right) \sin((2r+1)\Delta).
\end{align} 
Algorithm~\ref{alg:recovery} uses the above cutoff expansion to implement an approximate GD. We prove the convergence of this algorithm in two steps: (1) We first establish an approximation bound for the Fourier expansion, then (2) we prove that this approximation does not avoid the global convergence of subgradient descent when $m$ is sufficiently large. 
% For $\Delta = v - w$, we get 
% \begin{align}
%      g(v- \omega) = \sum_{r=0}^m \frac{4}{\pi(2r+1)}\left(\sin((2r+1)\theta) \cos((2r+1)\omega) - \cos((2r+1)\theta)\sin((2r+1)\omega)\right).      
% \end{align}
\subsection{Bounds for the Fourier series expansion}
\noindent Since the $\sign$ function is not continuous at zero, the Fourier series expansion is not point-wise convergence on $[-\Delta, \Delta]$. Indeed, Fourier expansion does not converge to the $\sign$ function at zero due to the \textit{Gibbs phenomena}~\citep{hewitt1979gibbs}. However, it is possible prove that $g$ converges to $\sign$ on $[-\Delta, \Delta]-\{0\}$ as $m\to \infty$.
\begin{lemma} \label{lemma:g_bounds}
$g$ obeys two important properties: 
\begin{itemize}
    \item[\textbf{I.}] $\forall \Delta \in \R- \{0\}$: \label{lemma:g_analysis}
    $
        | g(\Delta) - \sign(\Delta)| \leq  4 \left( \frac{1}{m |\Delta|} + \frac{1}{m}\right).
    $
    \item[\textbf{II.}] $\forall |\Delta|\leq \frac{\pi}{4}$ and $m>12$: $|g(\Delta)|\leq 1.9$.
\end{itemize}

\end{lemma}
\begin{proof}
  
We prove for $\Delta>0$, and is exactly the same for $\Delta<0$. 

\noindent
\textbf{I.}
According to the definition of~$g$, we have
\begin{align}
    | g(\Delta) - \sign(\Delta)| & = \left| \left(\frac{4}{\pi }\right) \underbrace{\sum_{r=m+1}^\infty  \frac{\sin((2r+1)\Delta)}{2r+1}}_{S_{m+1}(\Delta)} \right|.
    % & \leq  \left|\int_{2m \Delta}^\infty \left(\frac{ \sin(t) }{t}\right) dt \right| \\ 
    % & \leq  \left|\int_{0}^\infty \left(\frac{ \sin(2m\Delta)\cos(t)+\cos(2m\Delta)\sin(t) }{t}\right) dt \right|
\end{align} 
Let define $F_m$ as 
\begin{align}
    F_m(\Delta) = \sum_{r=1}^m \frac{\sin(r\Delta)}{r}.
\end{align}
Given $F_m(\Delta)$, $S_{m+1}(\Delta)$ reads as:
\begin{align} \label{eq:seq}
     S_{m+1}(\Delta)  = F_{\infty}(\Delta) -F_{2m}(\Delta)- \frac{1}{2}\left( F_{\infty}(2\Delta)- F_{m}(2 \Delta)\right).
\end{align}
To bound $S_m$, we first establish an integral form for $F_m(\Delta)$:
\begin{align} \label{eq:fm}
    F_m(\Delta) & = \sum_{r=1}^m \frac{\sin(r\Delta)}{r}  = \sum_{r=1}^m \int_0^{\Delta} \cos(rt) dt  =  \int_0^{\Delta} \sum_{r=1}^m  \cos(rt) dt.
\end{align}
Using Trigonometric identity $\sin(x) - \sin(y) = 2 \sin(x/2-y/2)\cos(x/2+y/2)$, we get
\begin{align}
   2 \sin(t/2) \sum_{r=1}^m \cos(rt) &= \sum_{r=1}^m \sin\left(\left(r+\frac{1}{2}\right)t\right) - \sin\left(\left(r-\frac{1}{2}\right)t\right) \\ 
   & = \sin\left( \left(m+\frac{1}{2}\right)t\right) - \sin\left(\frac{t}{2}\right).
\end{align}
Replacing the above identity into the expansion of $F_m$ yields 
\begin{align}
    F_m(\Delta) & = \int_0^{\Delta} \sum_{r=1}^m  \cos(rt) dt \\
    & = \frac{1}{2}\int_0^\Delta \frac{\sin\left( \left(m+\frac{1}{2} \right)t \right) }{\sin(\frac{t}{2})} dt - \frac{1}{2} \int_0^\Delta dt \\ 
    & =  \underbrace{\int_{0}^\Delta \left( \frac{1}{2\sin(t/2)} - \frac{1}{t} \right) \sin\left(\left(m+\frac{1}{2}\right)t\right)dt}_{\epsilon_m(\Delta)} + \int_{0}^{\Delta(m+\frac{1}{2})} \left(\frac{\sin(t)}{t}\right) dt - \frac{\Delta}{2}. \label{eq:fm_integral}
\end{align}
An application of integration by parts obtains
\begin{align} 
    |\epsilon_m(\Delta)| & \leq \left| \int_{0}^\Delta \left( \underbrace{\frac{1}{2\sin(t/2)}-\frac{1}{t}}_{h(t)}\right) \Im e^{(m+\frac{1}{2})t\Im} dt\right|\\
    & = \left| \frac{1}{m+\frac{1}{2}} \left( h(\Delta) e^{(m+\frac{1}{2})\Delta \Im} - \lim_{\epsilon \to 0} h(\epsilon) e^{\epsilon \Im}   + \int_{0}^\Delta  h'(t) e^{(m+\frac{1}{2})t\Im}dt \right) \right|  \\ 
    & \leq \frac{1}{m} \left( \int_{0}^\Delta  \underbrace{|h'(t)|}_{\leq 1/\pi} dt + \underbrace{|h(\pi)|}_{\leq 1/2} \right) \\
    & \leq \frac{3}{2m}. \label{eq:epsilonbound}
\end{align}
Replacing the above bound into Eq.\eqref{eq:seq} yields
\begin{align} \label{eq:smbound}
    |S_{m+1}(\Delta) | \leq \left| \int_{\Delta(2m+\frac{1}{2})}^\infty \frac{\sin(t)}{t} dt \right| + \frac{1}{2}\left| \int_{\Delta(2m+1)}^\infty \frac{\sin(t)}{t} dt \right| + \frac{3}{m}.
\end{align}
We use a change of variables and integration by parts to prove the following bound
\begin{align} 
    \left| \int_x^\infty \frac{\sin(t)}{t} dt \right| & = \left| \int_1^\infty \frac{\sin(xt)}{t} dt \right| \quad \quad \text{(change of variables)} \\ 
    & = \left| \int_1^\infty \frac{\Im \exp(\Im x t)}{t} dt \right| \\ 
    & = \frac{1}{x}\left| -e^{\Im  } +\int_1^{\infty} \frac{e^{\Im t}}{t^2} dt  \right| \quad \quad \text{(integration by parts)}\\ 
    & \leq \frac{2}{x}.
\end{align}
% Recall the following inverse Laplace transforms 
% \begin{align}
%     \frac{\cos(t)}{t} = \int_0^\infty \frac{s}{(1+s^2)} \exp(-t s) ds, \quad \frac{\sin(t)}{t} = \int_0^\infty \frac{1}{1+s^2} \exp(-ts)ds
% \end{align}
% An application of the Laplace transform obtains (see~\citep{mathover}):
% \begin{align}
%     \left|\int_{x}^\infty \frac{\sin(t)}{t} dt\right| & = \left|\int_{0}^\infty \left(\frac{ \sin(x)\cos(t)+\cos(x)\sin(t) }{t+x}\right) dt \right| \\ & =  \left|\int_{0}^\infty \left(\frac{ s\sin(x)+\cos(x) }{1+s^2}\right) \exp(- x s) ds \right| \\ 
%     & \leq \left| \int_{0}^\infty \exp(- x s) ds  \right| \leq \frac{1}{|x|} \label{eq:laplace_result}.
% \end{align}
Replacing this into Eq.~\ref{eq:smbound} concludes I. 

\noindent
\textbf{II.}
\noindent We write $g$ in terms of $F$ defined in Eq.~\eqref{eq:fm} as %\eqref{eq:epsilonbound}: 
\begin{align} 
    | g(\Delta) | & = \left|\frac{4}{\pi} \left(F_{2(m+1)}(\Delta) - \frac{1}{2} F_{m+1}(2\Delta)\right)\right|.
\end{align}
Eq.~\eqref{eq:fm_integral} allows us to decompose $g$ as
\begin{multline}
    \left|\frac{4}{\pi} \left(F_{2(m+1)}(\Delta) - \frac{1}{2} F_{m+1}(2\Delta)\right)\right|
    \\ \leq \frac{2}{\pi} \left| \int_{0}^{(2m+\frac{5}{2}) \Delta} \left(\frac{\sin(t)}{t}\right) dt \right| + \frac{4}{\pi} \left| \int_{(2m+\frac{5}{2}) \Delta}^{(2m+3)\Delta} \frac{\sin(t)}{t} dt \right| +  \frac{4}{\pi} \epsilon_{2m+1}(\Delta) + \frac{2}{\pi} \epsilon_{m}(2\Delta) \label{eq:gbound}.
\end{multline}
It is easy to check that the maximum of $h(x):= \int_0^x (\sin(t)/t) dt$ in $(0,\pi)$ is $h(\pi) \leq 1.86$:
\begin{align}
    \left|\int_{0}^x \frac{\sin(t)}{t} dt\right| & \leq \left|\int_{0}^\pi \frac{\sin(t)}{t} dt\right| = 1.86.
\end{align}
For $0<|\Delta|< \frac{\pi}{10}$, we get 
\begin{align}
     \left| \int_{(2m+\frac{5}{2}) \Delta}^{(2m+3)\Delta} \frac{\sin(t)}{t} dt \right|  \leq  \int_{(2m+\frac{5}{2}) \Delta}^{(2m+3)\Delta} \left|\frac{\sin(t)}{t}\right|  dt  \leq  \frac{\Delta}{2} \leq \frac{\pi}{20}
\end{align}
Using the established bound on $\epsilon_m$ in Eq.~\eqref{eq:epsilonbound}, we get
\begin{align}
    |g(\Delta)|\leq \underbrace{\frac{4\pi}{20 \pi} + \frac{2\times 1.86}{\pi}}_{\leq 1.5} +\left(\frac{4}{\pi}\right) \underbrace{\epsilon_{2m+1}(\Delta)}_{\leq \frac{3}{4m}} + \left(\frac{2}{\pi}\right) \underbrace{\epsilon_{m}(2\Delta)}_{\leq \frac{3}{2 m}}.
\end{align}
For $m\geq 12$, the above bound concludes
$
    |g(\Delta)| \leq 1.9
$.
\end{proof}
\subsection{Proof of Theorem~\ref{thm:finite_momments}}
 Since the objective $E$ and Algorithm~\ref{alg:recovery} are invariant to permutation of indices ${1,\dots, n}$, we assume $w_1 \leq  \dots\leq w_n$ without loss of generality. 
 Lemma~\ref{lemma:g_bounds} provides an upper-bound for the deviation of Algorithm~\ref{alg:recovery} from subgradient descent.
\noindent  Recall Algorithm~\ref{alg:recovery} obeys the following recurrence
\begin{align}
    v_{i}^{(k+1)} = v_i^{(k)} - \gamma \left(\frac{d \widehat{E}_k}{\| d \widehat{E}_k\|}\right), \quad d \widehat{E}_k := g(v_i^{(k)}-\omega_j)-\sum_{j} \sign(v_i^{(k)}-v_j^{(k)}).
\end{align}
\noindent
Let $\sigma$ is the optimal transport from $\nu^{(k)}$ to $\mu$. According to Lemma~\ref{lemma:optimal_transport}, $v_{\sigma(1)} \leq \dots \leq v_{\sigma(n)}$.
For the ease of notations, we use the compact notation $\Delta_j = v_{\sigma(j)}^{(k)} - \omega_{j}
$. 
The approximation error for the gradient is bounded as
\begin{align} 
      \left| \frac{d E}{d v_i}- d \widehat{E}_k\right| \leq \sum_{r}  | g(\Delta_r) - \sign(\Delta_r)| \leq  \begin{cases} 
      \frac{4n}{m} \left(1 + \frac{1}{\ell} \right)   & \forall j: |\Delta_j | \geq \ell  \\ 
      \frac{4n}{m} \left(1 + \frac{1}{\ell} \right)+ 1.9 & \text{otherwise,}
      \end{cases}
\end{align}
where we use the last Lemma to get the second inequality. Since $m \geq 800 n/\gamma$, we get
\begin{align} \label{eq:grad_bound} 
    \left| \frac{d E}{d v_i}- \widehat{g}_i \right| \leq  \begin{cases} 
      0.01 & \forall |\Delta_j | \geq \gamma  \\
      1.91 & \text{otherwise.}
      \end{cases}
\end{align}
The above inequality implies that the error for subgradient estimate is considerably small when $|\Delta_j\geq \gamma$.  It is easy to check that the above bound leads to 
$
    |v_{\sigma(i)}^{(k+1)} - w_i | \leq   |v_{\sigma(i)}^{(k)} - w_i | - \gamma
$
when $|\Delta_j| \geq \gamma$. However, Gibbs phenomena causes a large approximation error when $\Delta_i$ is small. \ref{assume:distinct}($\ell$) and $\gamma\leq \ell$ concludes that $|\Delta_j| \leq \gamma$ holds for at most one $j \in \{1, \dots, n\}$: 
\begin{itemize}
    \item[(a)] If $|\Delta_j| \leq \gamma$ for $j \neq \sigma(i)$, then the cardinality argument in Fig.~\ref{fig:elements_bound} leads to the following inequality:
    \begin{align}
        \left(\frac{d E}{d v_{\sigma(i)}}\right) \sign(v_{\sigma(i)} - w_i) \geq 2 \stackrel{\eqref{eq:grad_bound}}{\implies}    |v_{\sigma(i)}^{(k+1)} - w_i | \leq   |v_{\sigma(i)}^{(k)} - w_i | -  (0.09) \gamma
    \end{align}
    \item[(b)] 
If $|\Delta_{\sigma(i)}|\leq \gamma$, then the approximation bound in  Eq.~\eqref{eq:grad_bound} yields
$
  |v_{\sigma(i)}^{(k+1)} - w_i |  \leq 4 \gamma
$.
\end{itemize}
Putting all together, we get \begin{align}
    |v_{\sigma(i)}^{(k+1)} - w_i | \leq \begin{cases}
         |v_{\sigma(i)}^{(k)} - w_i | - (0.09)\gamma& |\Delta_{\sigma(i)}| \geq  \gamma \\ 
         3\gamma  & \text{otherwise}
    \end{cases}.
\end{align}
Induction over $k$ concludes the proof (similar to inductive argument in the proof of Lemma~\ref{lemma:particles}). 
\section{Neural networks}
\subsection{Proof of Proposition~\ref{lemma:nn}}
\citet{cho09deepkernel} prove 
\begin{align}
    \E \left[ \varphi(x^\top \widehat{w}) \varphi(x^\top \widehat{v})\right] = 1 - \frac{\theta(v,w)}{\pi}, \quad \theta(v,w) = \arccos(v^\top w)
\end{align}
holds for $x$ uniformly drawn from the unit circle and $v$ and $w$ on the unit circle. Plugging the polar coordinates into the above equation and incorporating the result in $L$ concludes the statement. 
\section{High-dimensional sparse measure recovery}

Now, we proof Lemma~\ref{lemma:high_dim} by casting a $d$-dimensional recovery to a set of one-dimensional instances, then leveraging Algorithm~\ref{alg:recovery}. Recall This algorithm can recover the individual coordinates according to Theorem~\ref{thm:finite_momments}. It remains to glue the coordinates. For the gluing, we need non-individual sensing of coordinates. We will show that the sensing of pair-wise sums of coordinates is sufficient to glue the coordinate and reconstruct the particles $w_1,\dots, w_n$. Algorithm~\ref{alg:recovery_d} the proposed algorithm. To analyze this algorithm, we rely on the following assumption the ensures the points have distinct coordinates. 

 \begin{algorithm}[t!]
\caption{Deterministic super-resolution} \label{alg:recovery_d}
\begin{algorithmic}
\Require  $\beta$ in \ref{assume:distinct_coordinates}, and $\epsilon$ \;
\For{$q=1,\dots,d$} 
 \State {Run Alg.~\ref{alg:recovery} with inputs $\Phi\mu=\frac{1}{n}\sum_{j=1}^n \Phi([w_j]_q)$ and stepsize $ \gamma=\min\{\frac{1}{2}\epsilon,\frac{1}{10}\beta\}$}
\State Store outputs $\{[v_i]_{q}\}_{i=1}^n$
\EndFor
\For{$q=1,\dots,d$}
\State Run Alg~\ref{alg:recovery} with inputs $\Phi\mu=\frac{1}{n}\sum_{j=1}^n \Phi([w_j]_q+[w_j]_1)$ and stepsize $ \gamma=\min\{\frac{1}{2}\epsilon,\frac{1}{10}\beta\}$ 

\State Store outputs
$\{[y_i]_{q}\}_{i=1}^n$ 
 \EndFor
\For{$i=1,\dots,n$}
\State $[v'_{i}]_1 \leftarrow [v_i]_1 $
\EndFor
\For{$i, j,r=1,\dots,n, q=2,\dots,d$}
\If{ $|[v_i]_1 + [v_j]_q - [y_r]_q| < \frac{1}{5}\beta $}
\State $[v'_{i}]_q \leftarrow [v_j]_q$
\EndIf
\EndFor
\State \textbf{Return} $\{ v'_1,\dots, v_{n}'\}$
\end{algorithmic}
\end{algorithm}
\begin{assumption}\label{assume:distinct_coordinates}
 Assume that there exists a positive constant $\beta>0$ such that
\begin{align*}
   0<\beta := \min \{  \min_{i\neq j,q} | [w_i]_q -[w_j]_q|, \min_{i\neq j,r>1} | [w_i]_1 + [w_i]_r - [w_j]_1 + [w_j]_r| \}. 
\end{align*}
\end{assumption}
Later, we will relax the above assumption to \ref{assume:distinct}. Yet, the above assumption simplifies our analysis for recovery guarantees presented in the next Lemma. 

\begin{lemma} \label{lemma:recovery_d} Suppose that \ref{assume:distinct_coordinates}($\beta$) holds; Algorithm~\ref{alg:recovery_d} returns $\{v'_i \in \R^d \}_{i=1}^n $  such that
\begin{align*}
   \max_{i}  \|v'_i - w_{\sigma(i)} \|_{\infty} \leq \epsilon
\end{align*}
holds for a permutation $\sigma$ of $\{1,\dots, n\}$ in $\bigo\left(n^2 d \left( \epsilon^{-2} +  \beta^{-2} + 1\right)\right)$ time. 
\end{lemma}
In other words,  Alg.~\ref{alg:recovery_d} solves super-resolution in polynomial-time under \ref{assume:distinct_coordinates}. Yet, \ref{assume:distinct_coordinates} is stronger than the standard \ref{assume:distinct}. The next lemma links these two assumptions with a random projection.

\begin{lemma} \label{lemma:assumptions}
Let $Z \in \R^{d\times d}$ be a random matrix with i.i.d.~Gaussian elements. Suppose points $w_1, \dots, w_n$ obey \ref{assume:distinct}($\ell$). With probability at least $1-\kappa$, points $\{ \frac{Zw_i}{\|Z\|}  \}_{i=1}^n$ obey \ref{assume:distinct_coordinates} with \[\beta =\Omega\left(\frac{\ell \kappa}{d n (-\ln(\kappa))}\right).\]
\end{lemma}
Thus, we can leverage a random projection to relax the strong assumption \ref{assume:distinct_coordinates} to the standard assumption \ref{assume:distinct}. We first use a random projection to make sure  \ref{assume:distinct_coordinates} holds; then, we use Algorithm~\ref{alg:recovery_d} to solve super-resolution under \ref{assume:distinct_coordinates}. Algorithm~\ref{alg:recovery_d_rand} presents the recovery method. 

\begin{algorithm}[h!]
\caption{Randomized deconvolution} \label{alg:recovery_d_rand}
\begin{algorithmic}
\Require The random matrix $Z \in \R^{d\times d}$, constant $\ell$ in \ref{assume:distinct}($\ell$), and $\kappa, \epsilon'>0$

\State Run Algorithm~\ref{alg:recovery_d} for points $w'_i := Z w_i/\|Z\|$ with $ \beta = \Omega\left(\frac{\ell \kappa}{d n (-\ln(\kappa))}\right)$ and $\epsilon = \frac{\epsilon'}{d}$
\State Store outputs $\{v'_i\}_{i=1}^n$ 

\State \textbf{Return} $\{ \|Z\|Z^{-1} v'_1,\dots, \|Z\|Z^{-1} v_{n}'\}$
\end{algorithmic}
\end{algorithm}

\begin{lemma}[Restated Lemma~\ref{lemma:high_dim}]\label{lemma:drecovery_weak}
Suppose $\{ w_i \in \S_{d-1} \}$ obey \ref{assume:distinct}($\ell$). 
 Algorithm~\ref{alg:recovery_d_rand} returns  $\{v_1, \dots, v_n \in \R^d \}$ such that there exits a permutation of $\{1, \dots, n\}$ denoted by $\sigma$ for which 
\begin{align*}
    \max_{i} \| v_i - w_{\sigma(i)}\| = \bigo( \epsilon) 
\end{align*}
holds with probability at least $1-\kappa$ for $\kappa \in (0,1)$. Furthermore, this algorithm has total computational complexity \[\bigo\left( (nd)^2 \epsilon^{-2} + (nd)^4 \left(\frac{\ln(\kappa)}{\kappa}\right)^2 \ell^{-2} + n d^3 \right).\]  
\end{lemma}
To the best of our knowledge, the above lemma establishes the first polynomial-time complexity for the recovery of measures over the unit sphere. Solvers based on convex relaxation rely on a root-finding algorithm~\citep{candes2014towards}. The complexity of this root-finding step is not studied when $w_i \in \R^d$. \citep{bredies2013inverse} develops an iterative method with the $\bigo(\text{iterations}^{-1})$ convergence rate. Yet,  each iteration is a non-convex program whose complexity has remained unknown.

\subsection{Proof of Lemma~\ref{lemma:recovery_d}}
Invoking Thm.~\ref{thm:finite_momments} yields: There exists a permutation $\sigma$ of indices $\{1,\dots,n\}$ such that
\begin{align}
    | [v_i]_q - [w_{\sigma(i)}]_q | \leq \min \{ \epsilon,\frac{1}{5}\beta \}
\end{align}
holds for all $i=1,\dots,n$. Similarly, there exists a permutation $\sigma'$ such that 
\begin{align}
| [y_{\sigma'(i)}]_q - \left( [w_{\sigma(i)}]_q + [w_{\sigma(i)}]_q \right) | \leq \epsilon
\end{align}
holds for all $i=1,\dots,n$. Using triangular inequality and  \ref{assume:distinct_coordinates}$(\alpha)$, we get 
\begin{align}
    |  [v_i]_q - [v_j]_q| & 
    \geq 
     | [v_i]_q - [v_j]_q \pm [w_{\sigma(i)}]_q \mp [w_{\sigma(j)}]_q| \\ 
    & \geq | [w_{\sigma(i)}]_q - [w_{\sigma(j)}]_q| - | [v_{\sigma(i)}]_q - [w_{\sigma(i)}]_q|  - | [v_{\sigma(j)}]_q - [w_{\sigma(j)}]_q|  \\ 
    & \geq | [w_{\sigma(i)}]_q - [w_{\sigma(j)}]_q| - 2\epsilon \geq \beta/2 \quad \quad \quad (\epsilon \leq \beta/4).
\end{align}
Hence
\begin{align}
     | [v_i]_1 + [v_i]_q - [y_{\sigma'(i)}]_q | & = | [v_i]_1 + [v_i]_q - [y_{\sigma'(i)}]_q \pm [w_{\sigma(i)}]_1 \mp  [w_{\sigma(i)}]_q  | \\
     & \leq | [v_i]_1 - [w_{\sigma(i)}]_1| +| [v_i]_q - [w_{\sigma(i)}]_q| + | [y_{\sigma'(i)}]_q - \left( [w_{\sigma(i)}]_q + [w_{\sigma(i)}]_q \right) | \\ &  \leq 3\epsilon. \label{eq:bound_3eps}
\end{align}
Furthermore for all $j \neq i$, the following holds 
\begin{align}
     | [v_i]_1 + [v_i]_q - [y_{\sigma'(j)}]_q | & \geq | [v_i]_1 + [v_i]_q - [y_{\sigma'(j)}]_q \mp [y_{\sigma'(i)}]_q \pm [w_{\sigma(i)}]_1  \mp  [w_{\sigma(i)}]_q  | \\
     & \geq |[y_{\sigma'(j)}]_q - [y_{\sigma'(i)}]_q | -  | [v_i]_1 + [v_i]_q - [y_{\sigma'(i)}]_q \pm [w_{\sigma(i)}]_1 \mp  [w_{\sigma(i)}]_q  |  \\
     & \geq \beta - 3\epsilon \geq \beta/5, \quad \quad \quad \quad \text{(using Eq.\ref{eq:bound_3eps})} . 
\end{align}
Combining the last two inequalities ensures that the gluing in Alg.~\ref{alg:recovery} obtains an $\epsilon$-accurate solution in norm-$\infty$. To complete the statement proof, we analyze the complexity of Algorithm~\ref{alg:recovery_d}. The algorithm has two main parts: (i) the recovery of individual coordinates, and (ii) gluing coordinates. 
\begin{itemize}
    \item[(i)] Theorem~\ref{thm:finite_momments} establishes $\bigo(n^2(\epsilon^{-2}+ \beta^{-2}))$ complexity for Alg.~\ref{alg:recovery}. Alg.~\ref{alg:recovery_d} recall this Alg.~\ref{alg:recovery} $\bigo(d)$ times which costs $\bigo(dn^2(\epsilon^{-2}+ \beta^{-2}))$.
    \item[(ii)] The gluing step is $\bigo(n^2 d)$.
\end{itemize}
Thus, the total complexity is $\bigo(dn^2(\epsilon^{-2}+ \beta^{-2}+1))$.

\subsection{Proof of Lemma~\ref{lemma:assumptions}}

Let $\Delta_{ij} := w_i - w_j$, and $z_q \in \R^d$ denote the rows of $Z$. Then, 
$
     \zeta_q(ij) = (z^\top_q \Delta_{ij})^2/\| \Delta_{ij} \|^2   
$ is a $\chi$-square random variable for which the following bound holds 
\begin{align}
     P\left( \| \Delta_{ij} \|^2 \zeta_q(ij) \leq a^2 \|\Delta_{ij}\|^2  \right) \leq  a. 
\end{align}
Setting $a =\kappa \ell/(2\sqrt{d}n)$ and union bound over all $i,j,$ and $q$ concludes 
\begin{align}
 P\left( \left\{\exists i \neq j,q: |z^\top_{q} (w_i-w_j)|\leq a   \right\}\right) \leq \kappa/2.
\end{align}
Thus
\begin{align}
    |(z_1+z_{r})^\top (w_i-w_j)| \leq | z_1^\top (w_i-w_j) | + | z_r^\top (w_i-w_j)| \leq 2a
\end{align}
holds with probability $1-\kappa/2$. To complete the proof, we invoke spectral concentration bounds for random matrices presented in the next Lemma. 
\begin{lemma}[ \citep{rudelson2010non}] \label{lemma:norm} 
For matrix $Z$ with i.i.d.~standard normal elements, there exists constants $c$ and $C$ such that 
\begin{align}
    P( \| Z \| \geq (2+a)\sqrt{d}) \leq C\exp(- c d a^{3/2})
\end{align}
holds. 
\end{lemma}

According to the above Lemma, the spectral norm of the random matrix $Z$ is $\Omega((-\ln(\kappa)+2) \sqrt{d})$ with probability $1- \kappa/2$. Combining the spectral bound and the last inequality completes the proof.  

\subsection{Proof of Lemma~\ref{lemma:drecovery_weak}}
We will use the following concentration bound for the smallest eigenvalues of the random matrix $Z$. 
\begin{lemma}[Theorem 1.2 of \citep{szarek1991condition}] \label{lemma:condition}
There is an absolute constant $c$ such that 
\begin{align}
     P\left( \| Z^{-1} \| \geq  c\sqrt{d}/\kappa \right) & \leq \kappa.
\end{align}
\end{lemma}
\noindent According to Lemma~\ref{lemma:recovery_d}, Algorithm~\ref{alg:recovery_d} returns $\{ v'_i \}_{i=1}^n$ for which
\begin{align}
      \max_i \|v'_i - w'_{\sigma(i)} \|_{\infty} \leq  \frac{\epsilon}{d}
\end{align}
holds with probability $1-\kappa$.
Recall the output of Algorithm~\ref{alg:recovery_d_rand} :$v_i = \|Z \| Z^{-1} v'_i$. Using the above bound, we complete the proof:
\begin{align}
    \| v_i - w_{\sigma(i)}\| & =  \|\| Z\| Z^{-1} v'_i - \|Z \| Z^{-1} \underbrace{ \left(Z w_{\sigma(i)}/\|Z\|\right)}_{w'_{\sigma(i)}} \| \\
    & \leq \|Z\|\| Z^{-1} \| \|v'_i - w'_{\sigma_i}\| \\ 
    & \leq \bigo(d)  \|v'_i - w'_{\sigma_i}\|_\infty, \quad \quad \text{(Lemmas~\ref{lemma:condition}, and \ref{lemma:norm})}\\
    & \leq \bigo(\epsilon).
\end{align}
Finally, we calculate the complexity using Lemma~\ref{lemma:recovery_d}.  Replacing $\beta$ and $\epsilon/d$ in complexity of Algorithm~\ref{alg:recovery_d} leads to
\begin{align}
 \bigo\left( (nd)^2 \epsilon^{-2} + (nd)^4 \left(\frac{\ln(\kappa)}{\kappa}\right)^2 \ell^{-2}  \right)   
\end{align}
computational complexity. In addition to the above complexity, the matrix multiplications with $Z$ and $Z^{-1}$
takes $O(n d^3)$, which concludes the complexity of Algorithm~\ref{alg:recovery_d_rand}.

\end{document}